\documentclass[twoside]{article}
\usepackage[accepted]{aistats2015}

\usepackage{hyperref}
\usepackage{url}

\usepackage{wrapfig}
\usepackage{amssymb} 
\usepackage{graphicx}
\usepackage{amsmath}
\usepackage{amsthm}
\usepackage{subfigure}
\usepackage[latin1]{inputenc}
\usepackage{color}
\usepackage{multirow}
\usepackage[hypcap=false]{caption}
\usepackage{rotating}
\usepackage{multirow}
\usepackage{picins}

\usepackage{algorithm}
\usepackage{algpseudocode}

\usepackage{booktabs}
\newtheorem{prop}{Proposition}

\allowdisplaybreaks

\usepackage{bm}
\usepackage{xspace}
\usepackage{amssymb}
\usepackage{amsmath}
\usepackage{amsfonts}
\usepackage{euscript}

\newcommand{\bbZero}{{\bf 0}}

\newcommand{\mD}{{\mathcal D}}

\newcommand{\mN}{{\mathcal N}}

\newcommand{\mG}{{\mathcal G}}
\newcommand{\mO}{{\mathcal O}}

\newcommand{\mbbR}{{\mathbb R}}

\newcommand{\mbbE}{{\mathbb E}}

\def\registered{{\ooalign {\hfil\raise .05ex\hbox{\scriptsize
R}\hfil\crcr\mathhexbox20D}}}

\def\REgistered{{\ooalign
{\hfil\raise.09ex\hbox{\tiny \sf R}\hfil\crcr\mathhexbox20D}}}

\makeatletter
\DeclareRobustCommand\onedot{\futurelet\@let@token\@onedot}
\def\@onedot{\ifx\@let@token.\else.\null\fi\xspace}

\graphicspath{{pics/}{figs/}}

\newcommand{\manSamp}{0.23\textwidth}

\newcommand{\embeddings}{0.2\textwidth} %

\begin{document}

\runningtitle{Sampling from DPPs for Scalable Manifold Learning}

\twocolumn[

\aistatstitle{Diverse Landmark Sampling from Determinantal Point Processes \\ for Scalable Manifold Learning}

\aistatsauthor{ Christian Wachinger \And Polina Golland }

\aistatsaddress{ CSAIL, MIT \\ MGH, Harvard Medical School \And CSAIL, MIT  } ]

\begin{abstract} 
High computational costs of manifold learning prohibit its application for large point sets. 
A common strategy to overcome this problem is to perform dimensionality reduction on selected landmarks and to successively embed the entire dataset with the Nyström method.
The two main challenges that arise are: 
(i)~the landmarks selected in non-Euclidean geometries must result in a low reconstruction error, 
(ii) the graph constructed from sparsely sampled landmarks must approximate the manifold well. 
We propose the sampling of landmarks from determinantal distributions on non-Euclidean spaces. 
Since current determinantal sampling algorithms have the same complexity as those for manifold learning, we present an efficient approximation running in linear time.  
Further, we recover the local geometry after the sparsification by assigning each landmark a local covariance matrix, estimated from the original point set. 
The resulting neighborhood selection based on the Bhattacharyya distance improves the embedding of sparsely sampled manifolds. 
Our experiments show a significant performance improvement compared to state-of-the-art landmark selection techniques. 
\end{abstract}

\section{Introduction}
Spectral methods are central for a multitude of applications in machine learning, statistics, and computer vision, such as dimensionality reduction~\cite{tenenbaum2000global,wachinger2012manifold}, classification~\cite{reuter:cad06,wachinger2015brainprint}, and segmentation~\cite{shi2000normalized,wachinger12spectral}. 
A limiting factor for the spectral analysis is the computational cost of the eigen decomposition. %
To overcome this limitation, the Nyström method~\cite{williams2001using} is commonly applied to approximate the spectral decomposition of the Gramian matrix. 
A subset of rows/columns is selected and based on the eigen decomposition of the resulting small sub-matrix, the spectrum of the original matrix can be approximated. 
While the Nyström extension is the standard method for the matrix reconstruction, the crucial part is the subset selection. 
In early work~\cite{williams2001using}, uniform sampling without replacement was proposed. 
This was followed by numerous alternatives including K-means clustering~\cite{zhang2008improved}, greedy approaches~\cite{ouimet2005greedy,farahat2011novel}
, and volume sampling~\cite{belabbas2009landmark,deshpande2006matrix}. 
A recent comparison of several approaches is presented in~\cite{Kumar12}. 

Of particular interest for subset selection is volume sampling~\cite{deshpande2006matrix}, 
equivalent to determinantal sampling~\cite{belabbas2009landmark}, 
because reconstruction error bounds exist. 
This method is, however, not used in practice because of the high computational complexity of sampling from the underlying distributions~\cite{Kumar12}. 
Independently, determinantal point processes (DPPs) have been proposed recently for tracking and pose estimation~\cite{kulesza2010structured}. 
They were originally designed to model the repulsive interaction between particles. 
DPPs are well suited for modeling diversity in a point set. 
A sampling algorithm for DPPs was presented in~\cite{hough2006determinantal,kulesza2010structured}, which has complexity $\mO(n^3)$ for $n$ points. 
Since this algorithm has the same complexity as the spectral analysis, it cannot be directly used as a subset selection scheme.

In this paper, we focus on nonlinear dimensionality reduction for large datasets via manifold learning. 
Popular manifold learning techniques include kernel PCA~\cite{scholkopf1998nonlinear}, Isomap~\cite{tenenbaum2000global}, and Laplacian eigenmaps~\cite{belkin03}. 
All of these methods are based on a kernel matrix of size $\mO(n^2)$ that contains the information about the pairwise relationships between the input points. 
The spectral decomposition of the kernel matrix leads to the low-dimensional embedding of the points. 
For large $n$, one is interested in avoiding its explicit calculation and storage. 
In contrast to general rank-$k$ matrix approximation, 
this is possible by taking the nature of the non-linear dimensionality reduction into account and relating the entries of the kernel matrix directly to the original point set. 

Consequently, we propose to perform DPP sampling on the original point set to extract a diverse set of landmarks. 
Since the input points lie in a non-Euclidean space, ignoring the underlying geometry leads to poor results. 
To account for the non-Euclidean geometry of the input space, we replace the Euclidean distance with the geodesic distance along the manifold, which is approximated by the graph shortest path distance. 
Due to the high complexity of DPP sampling, we derive an efficient approximation that runs in $\mO(ndk)$, with input dimensionality~$d$ and subset cardinality~$k$. 
The algorithm restricts the updates to be local, which enables sampling on complex geometries. 
This, together with its low computational complexity, makes the algorithm well suited for the subset selection in large scale manifold learning. 

A consequence of the landmark selection is that the manifold is less densely sampled than before, making its approximation with neighborhood graphs more difficult. 
It was noted in~\cite{Balasubramanian04012002}, as a critical response to~\cite{tenenbaum2000global}, that the approximation of manifolds with graphs is topologically unstable. 
In order to improve the graph construction, we retain the local geometry around each landmark by locally estimating the covariance matrix on the original point set. 
This allows us to compare multivariate Gaussian distributions with the Bhattacharyya distance for neighborhood selection, yielding improved embeddings.
A shorter version of this work was published in~\cite{wachigner15dpp}.

\section{Background}
We assume $n$ points in high dimensional space $x_1, \ldots, x_n \in \mbbR^d$ and let $X \in \mbbR^{d\times n}$ be the matrix whose $i$-th column is the point $x_i$. 
Non-linear dimensionality reduction techniques are based on a positive semidefinite kernel $K$, with a typical choice of Gaussian or heat kernel $K_{i,j} = \exp(- \| x_i - x_j \|^2 / 2 \sigma^2)$. The resulting kernel matrix is of size $\mO(n^2)$. 
Necessary for spectral analysis is the eigen decomposition of the kernel matrix, which has complexity $\mO(n^3)$. 
For most techniques, it is only necessary to compute the leading $k$ eigenvectors. 
The problem can therefore also be considered as finding the best rank-$k$ approximation of the matrix $K$, with the optimal solution $K_k = \sum_{i=1}^k \lambda_i u_i u_i^\top$, where $\lambda_i$ is the $i$-th largest eigenvalue and $u_i$ is the corresponding eigenvector.

\subsection{Nyström Method}
Suppose  $J \subseteq \{1, \ldots, n\}$ is a subset of the original point set of size $k$ and $\bar{J}$ is its complement. 
We can reorder the kernel matrix $K$ such that
{\small
\begin{equation*}
K = \begin{bmatrix} K_{J \times J} & K_{J \times \bar{J}} \\ K_{J \times \bar{J}}^\top & K_{\bar{J} \times \bar{J}} \end{bmatrix},
\ \ \ \tilde{K} = \begin{bmatrix} K_{J \times J} & K_{J \times \bar{J}} \\ K_{J \times \bar{J}}^\top & K_{J \times \bar{J}}^\top K^{-1}_{J \times J} K_{J \times \bar{J}} \end{bmatrix} 
\end{equation*}}
\normalsize 
\noindent
with $\tilde{K}$ being the matrix estimated via the Nyström method~\cite{williams2001using}. 
The Nyström extension leads to the approximation $K_{\bar{J} \times \bar{J}} \approx K_{J \times \bar{J}}^\top K^{-1}_{J \times J} K_{J \times \bar{J}}$. 
The matrix inverse is replaced by the Moore-Penrose generalized inverse in case of rank deficiency. 
The Nyström method leads to the minimal kernel completion~\cite{belabbas2009landmark} conditioned on the selected landmarks and has been reported to perform well in numerous applications~\cite{chen2010parallel,fowlkes2004spectral,Platt05fastmap,Talwalkar08,williams2001using}. 
The challenge lies in finding landmarks that minimize the reconstruction error 
\begin{equation}
\| K - \tilde{K} \|_{\text{tr}} = \text{tr}(K_{\bar{J} \times \bar{J}}) - \text{tr}(K_{J \times \bar{J}}^\top K^{-1}_{J \times J} K_{J \times \bar{J}}).
\label{equ:error}
\end{equation}
The trace norm $\|.\|_{\text{tr}}$ is applied because results only depend on the spectrum due to its unitary invariance. 

\subsection{Annealed Determinantal Sampling\label{sec:aDetSamp}}
A large variety of methods have been proposed for selecting the subset $J$. 
For general matrix approximation, this step is referred to as row/column selection of the matrix $K$, 
which is equivalent to selecting a subset of points $X$. %
This property is important because it avoids explicit computation of the $\mO(n^2)$ entries in the kernel matrix $K$. 
We focus on volume sampling for subset selection because of its theoretical advantages~\cite{deshpande2006matrix}.  
We employ the factorization $K_{J \times J} = Y^\top_J Y_J$, which exists because $K_J$ is positive semidefinite. 
Based on this factorization, the volume $\text{Vol}(\{ Y_i \}_{i \in J})$ of the simplex spanned by the origin and the selected points $Y_J$ is calculated, which is equivalent to the volume of the parallelepiped spanned by $Y_J$.  
The subset $J$ is then sampled proportional to the squared volume. 
This is directly related to the calculation of the determinant with $\det(K_{J \times J}) = \det(Y^\top_J Y_J) = \det(Y_J)^2 = \text{Vol}^2( \{ Y_i \}_{i \in J})$. 
These ideas were further generalized in~\cite{belabbas2009landmark} based on annealed determinantal distributions
\begin{equation}
p^s(J) \propto \det(K_{J \times J})^s = \det(Y_J^\top Y_J)^s = \det(Y_J)^{2s}.
\end{equation}
This distribution is well defined because the principal submatrices of a positive semidefinite matrix are themselves positive semidefinite. 
Varying the exponent $s \geq 0$ results in a family of distributions, modeling the annealing behavior as used in stochastic computations. 
For $ s = 0 $ this is equivalent to uniform sampling~\cite{williams2001using}.  
In the following derivations, we focus on $s=1$. 
It was shown in~\cite{deshpande2006matrix} that for $J \sim p(J), |J|=k$ 
\begin{equation}
{\mathbb E}  \left[ \| K - \tilde{K}\|_{\text{tr}} \right] \leq (k+1) \| K - K_k \|^2_F, 
\end{equation} 
where $\tilde{K}$ is the Nyström reconstruction of the kernel based on the subset $J$, $K_k$ the best rank-$k$ approximation achieved by selecting the largest eigenvectors, and $\|.\|_F$ the Frobenius norm. 
It was further shown that the factor $k+1$ is the best possible for a $k$-subset. 
Related bounds were presented in~\cite{belabbas2009spectral}. 
A further result is that for any matrix $K$, there exist $k+k(k+1)/\epsilon$ rows whose span contains the rows of a rank-$k$ matrix $\tilde{K}_k$ such that  
\begin{equation}
\| K - \tilde{K}_k \|_F^2 \leq (1+\varepsilon) \| K - K_k \|_F^2.
\end{equation}
This result establishes a connection between matrix approximation and projective clustering, with the selection of a subsets columns being similar to the construction of a coreset~\cite{agarwal2005geometric,FeldmanFK11}.

\begin{figure*}
\begin{center}
\subfigure[Swiss Roll\label{fig:roll}] {
	\includegraphics[width=\manSamp]{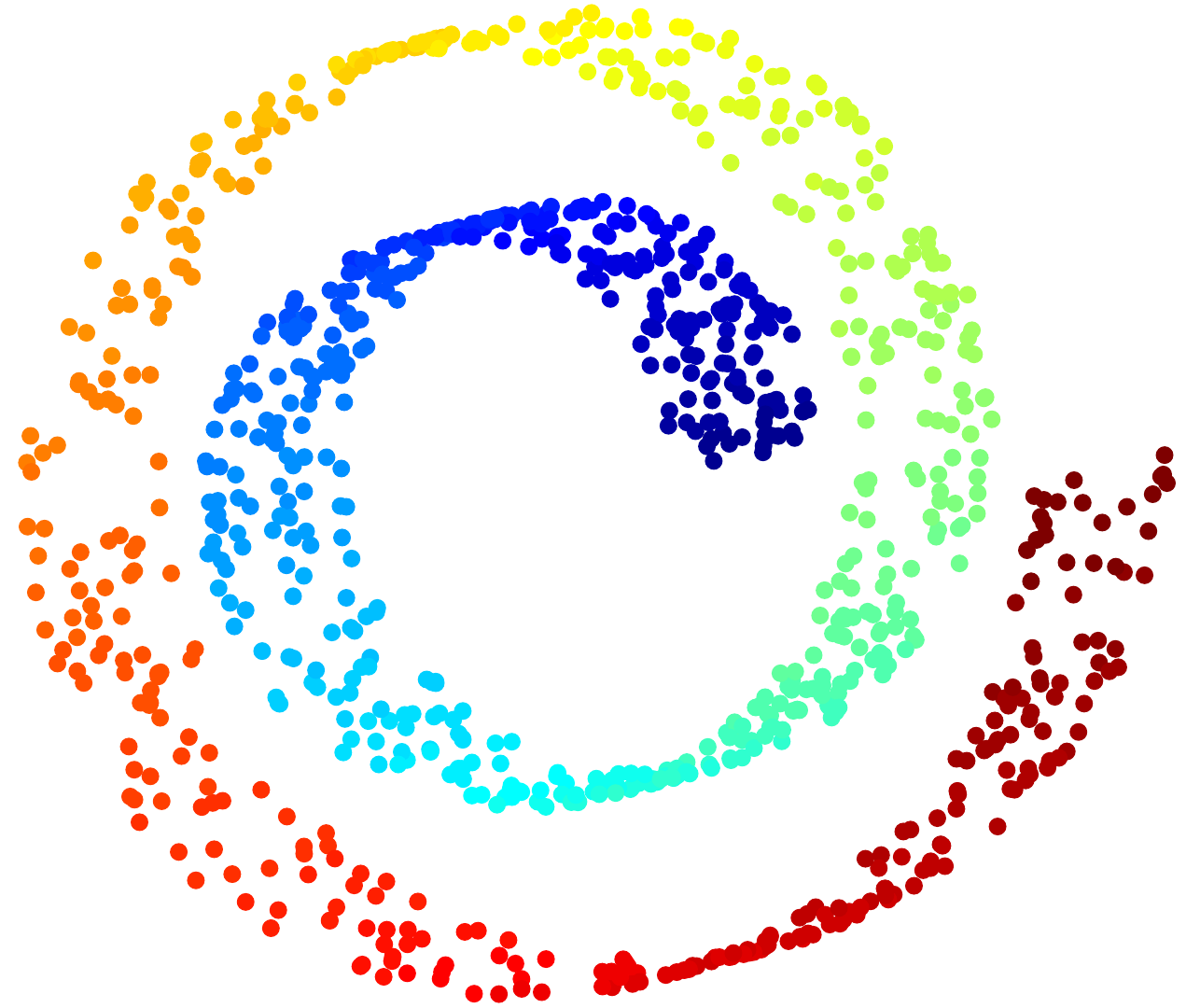} 
}
\subfigure[Standard\label{fig:rollReg}] {
	\includegraphics[width=\manSamp]{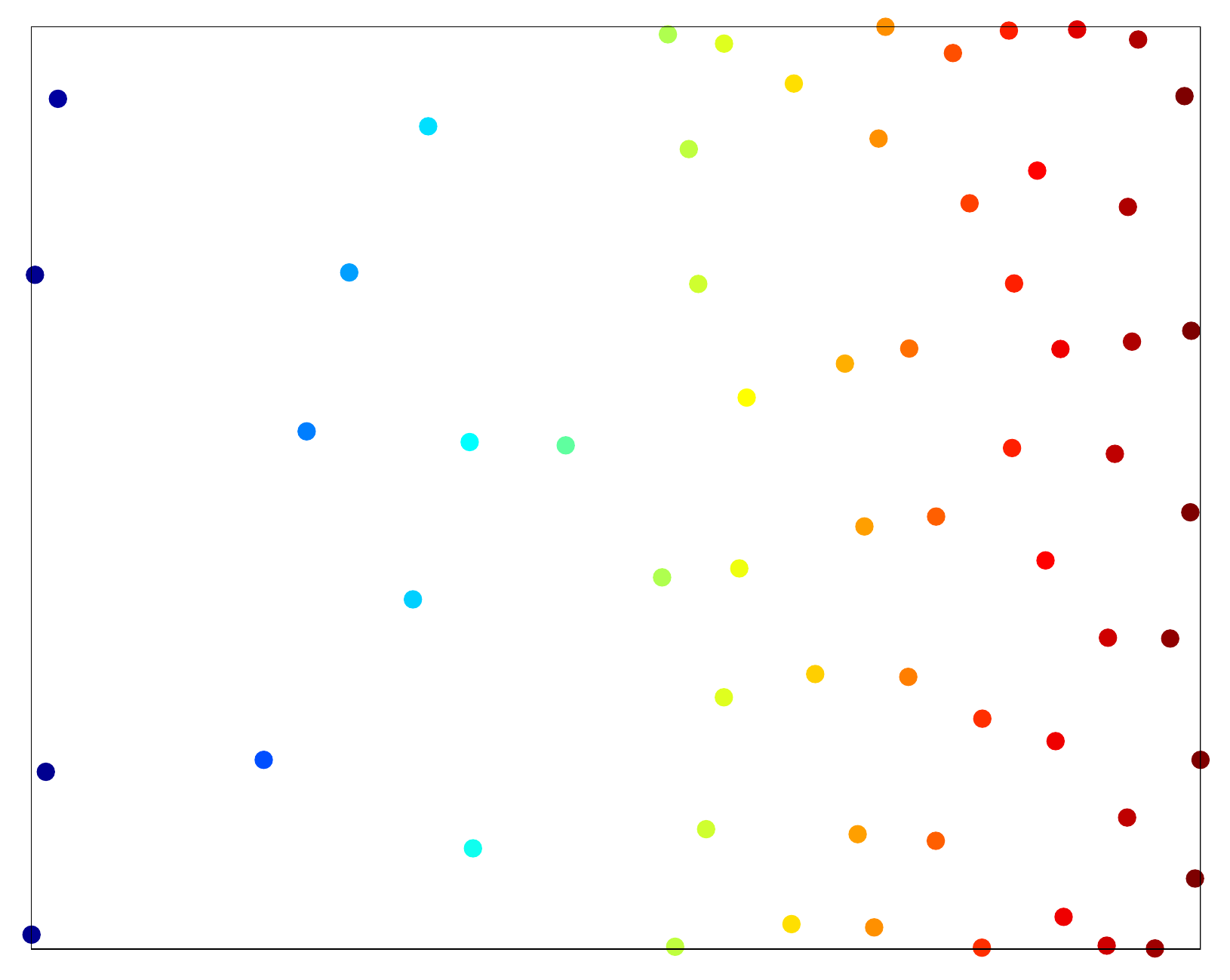} 
}
\subfigure[Geodesic\label{fig:rollGraph}] {
	\includegraphics[width=\manSamp]{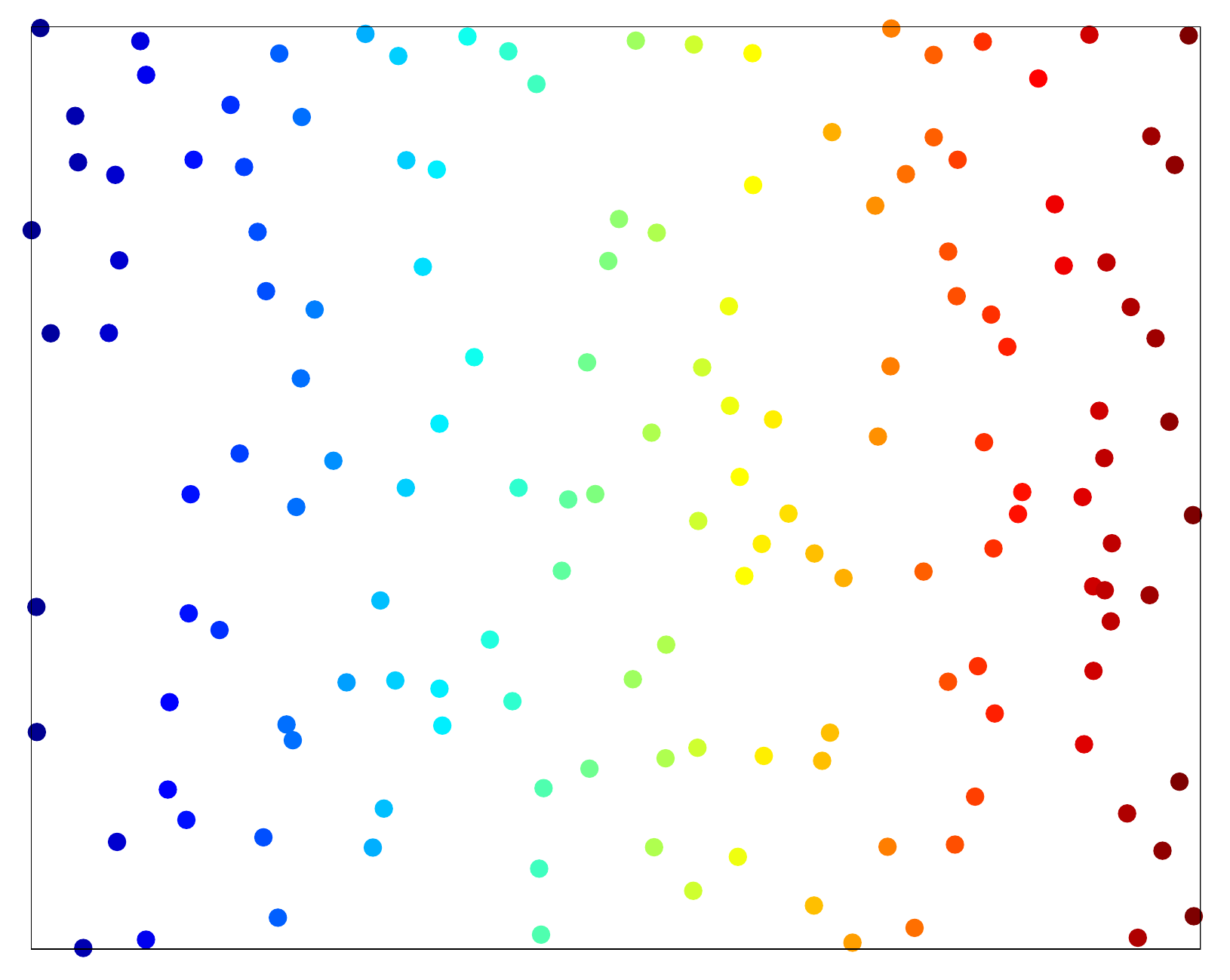} 
}
\subfigure[Efficient\label{fig:rollApp}] {
	\includegraphics[width=\manSamp]{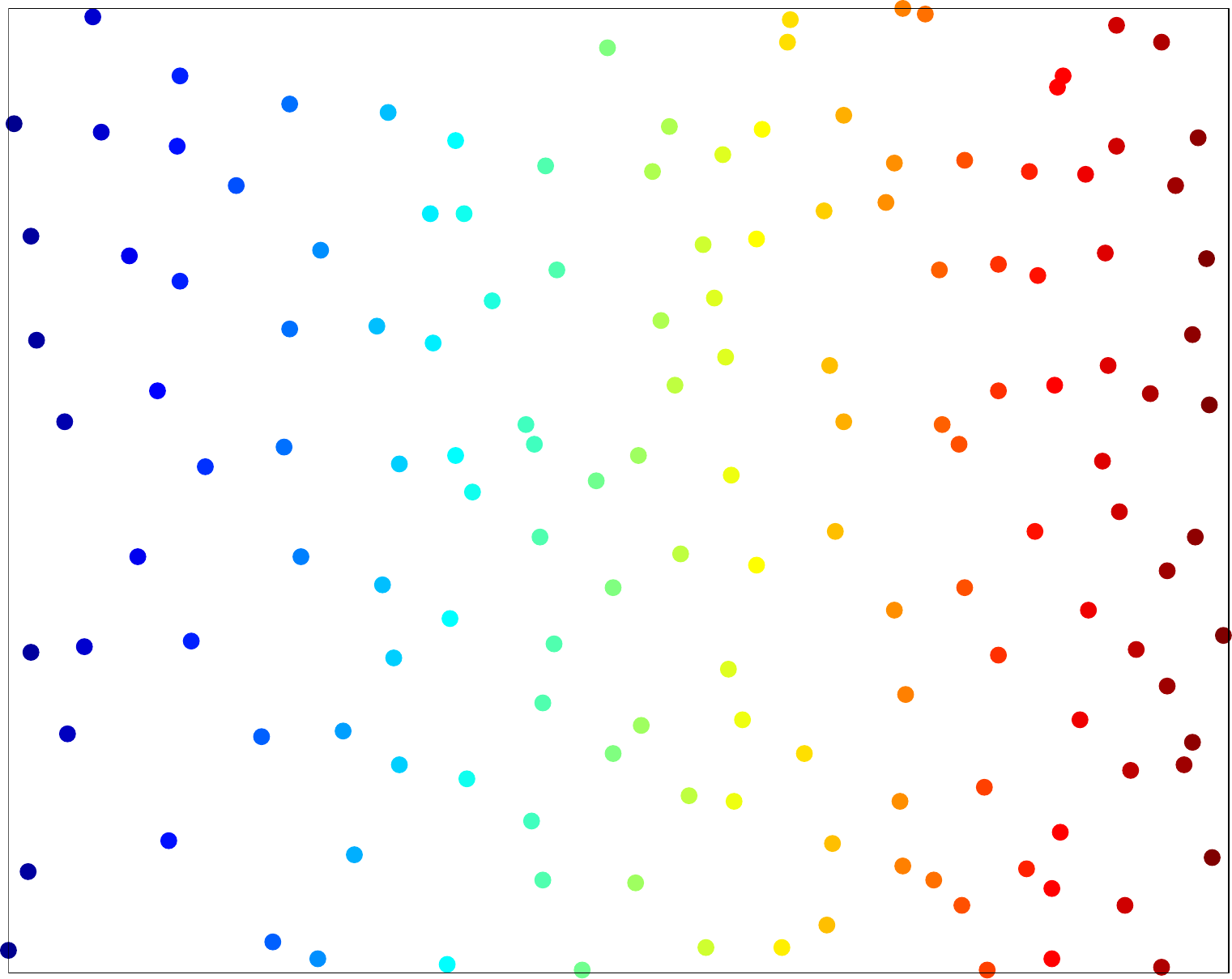} 
}
\caption{DPP sampling from 1,000 points lying on a manifold. We show results for standard DPP sampling, geodesic DPP sampling, and efficient DPP sampling. 
Note that the sampling is performed in 3D, but we can plot the underlying 2D manifold by reversing the construction of the Swiss roll. Geodesic and efficient sampling yields a diverse subset from the manifold.
\label{fig:boxplot}}
\end{center}
\end{figure*}

\section{Method}

In the following, we will first analyze the sampling from determinantal distributions on  non-Euclidean geometries. 
Subsequently, we introduce an efficient algorithm for approximate DPP sampling on manifolds. 
Finally, we present our approach for robust graph construction on sparsely sampled manifolds. %

\subsection{DPP Sampling on Manifolds\label{sec:manSamp}}
As described in Section~\ref{sec:aDetSamp}, sampling from determinantal distributions is used for row/column selection. 
Independently, determinantal point processes (DPPs) were introduced for modeling probabilistic mutual exclusion~\cite{macchi1975coincidence}. 
They present an attractive scheme for ensuring diversity in the selected subset.  
An interesting construction of DPPs is based on L-ensembles~\cite{Borodin09}. 
Given a positive semidefinite matrix $L \in \mbbR^{n \times n}$,  the likelihood for selecting the subset $J \subseteq \{1, \ldots, n\}$ is 
\begin{equation}
P_L(J) = \frac{\det(L_{J \times J})}{\det(L + I)},
\end{equation}
where $I$ is the identity matrix and $L_{J \times J}$ is the sub-matrix of $L$ containing the rows and columns indexed by $J$. 
Identifying the L-ensemble matrix $L$ with the kernel matrix $K$, we can apply DPPs to sample subsets from the point set $X$.

To date, applications using determinantal point processes have assumed Euclidean geometry~\cite{borodin2010adding,kulesza2010structured}. 
For non-linear dimensionality reduction, we assume that the data points lie on non-Euclidean spaces, such as the Swiss roll in Fig.~\ref{fig:roll}. 
To evaluate the performance of DPPs on manifolds, we sample from the Swiss roll. 
Since we know the construction rule in this case, we can invert it and display the sampled 3D points in the underlying 2D space. 
The result in Fig.~\ref{fig:rollReg} shows that the inner part of the roll is almost entirely neglected, as a consequence of not taking the manifold structure into account. 
A common solution is to use geodesic distances~\cite{tenenbaum2000global}, which can be approximated by the graph shortest path algorithm. 
Consequently, we replace the Euclidean distance~$\| .\|$ in the construction of the kernel matrix $K$ with the geodesic distance $K_{i,j} = \exp( - \|x_i - x_j\|_{\text{geo}}^2/2\sigma^2)$. 
The result is shown in Fig.~\ref{fig:rollGraph}. 
We observe a clear improvement in the diversity of the sampling, now also including points in the interior part of the Swiss roll.

\subsection{Efficient Approximation of DPP Sampling on Manifolds}
We have seen in the last sections that it is possible to adapt determinantal sampling to non-Euclidean geometries and that error characterizations for the subset selection exist. 
However, we are missing an efficient sampling algorithm for dealing with large point sets. 
In~\cite{deshpande2006matrix}, an approximative sampling based on the Markov chain Monte Carlo method is proposed to circumvent the combinatorial problem with ${n\choose k}$ possible subsets. 
Further approximations include sampling proportional to the diagonal elements $K_{ii}$ or its squared version $K_{ii}^2$, leading to additive error bounds~\cite{belabbas2009spectral,Drineas05onthe}. 
In~\cite{deshpande2006adaptive}, an algorithm is proposed that yields a $k!$ approximation to volume sampling, worsening the approximation from $(k+1)$ to $(k+1)!$.

\begin{algorithm}
\caption{DPP sampling equivalent to~\cite{kulesza2010structured} \label{alg:dpp}}
\begin{algorithmic}[1]
\Require Eigen decomposition of $K$: $\{(v_i, \lambda_i)\}_{i=1}^n$ 
\For{i = 1 to n}
	\State Add eigenvector $v_i$ with probability $\frac{\lambda_i}{\lambda_i +1}$ to $V$
\EndFor
\State $B = V^\top$
\For{1 to $|V|$}
	\State Select $i \in 1 \ldots n$ with probability $P(i) \propto \|B_i\|^2$ 
	\State $J \gets	J \cup i$
	\State $B_j \gets \text{Proj}_{\perp B_i} B_j$  for all  $j \in \{1, \ldots, n\}$ 
\EndFor
\State \Return $J$
\end{algorithmic}
\end{algorithm}

An exact sampling algorithm for DPPs was presented in~\cite{hough2006determinantal,kulesza2010structured}, which requires the eigen decomposition of $K = \sum_{i=1}^n \lambda_i v_i v_i^\top$. 
We state an equivalent reformulation of this sampling algorithm in Algorithm~\ref{alg:dpp}. 
First, eigenvectors are selected proportional to the magnitude of their eigenvalues and stored as columns in $V$. 
Assuming $m$ vectors are selected then $V \in \mbbR^{n \times m}$. 
By setting $B = V^\top$, we denote the rows of $V$ as $B_i \in \mbbR^m$. 
In each iteration, we select one of the $n$ points where point $i$ is selected proportional to the squared norm $\|B_i\|^2$. 
The point is added to the subset $J$. %
After the selection of~$i$, all vectors $B_j$ are projected to the orthogonal space of $B_i$. 
Since $\text{Proj}_{\perp B_i} B_i = 0$, the same point is almost surely not selected twice. 
The update formulation differs from~\cite{kulesza2010structured}, where an orthonormal basis of the eigenvectors in $V$ to the $i$-th basis vector $e_i \in \mbbR^n$  is constructed. 
Both formulations are equivalent but provide a different point of view on the algorithm. 
The modification is, however, essential for the justification of the proposed efficient sampling procedure. 
The following proposition characterizes the behavior of the update in the algorithm. %
\begin{prop}
\label{thm1}
Let $B_i, B_j \in \mbbR^m \setminus \{\bbZero \}$ be two non-zero vectors in $\mbbR^m$, and $\theta = \angle(B_i,B_j)$ be the angle between them. 
Then
\begin{equation}
\|\text{Proj}_{\perp B_i} B_j \|^2 =  \| B_j \|^2  \sin^2 \theta, 
\end{equation}
where $\text{Proj}_{\perp B_i} B_j$ is the the projection of $B_j$ on the subspace perpendicular to $B_i$. 
For  $B_i \neq \bbZero$ and $B_j = \bbZero$ the projection is $\|\text{Proj}_{\perp B_i} B_j \|^2 = 0$. 
\end{prop}
We state the proof in the supplementary material.

Sampling from a determinantal distribution is not only advantageous because of the presented error bounds but it also makes intuitive sense that a selection of a diverse set of points yields an accurate matrix reconstruction.
The computational complexity of Algorithm~\ref{alg:dpp} is, however, similar or even higher than for manifold learning because the spectral decomposition of a dense graph is required, whereas Laplacian eigenmaps operate on sparse matrices.   
An approach for efficient sampling proposed in~\cite{kulesza2010structured} works with the dual representation of $K = Y^\top Y$ to obtain $Q = Y Y^\top$, with $Q$ being hopefully smaller than the matrix $K$. 
Considering that we work with a Gaussian kernel matrix, this factorization corresponds to the inner product in feature space $\phi(x_i)^\top \phi(x_j)$ of the original points $x_i, x_j$. 
It is noted in the literature that the Gaussian kernel corresponds to an infinite dimensional feature space~\cite{scholkopf1998nonlinear}. 
Since we work with symmetric, positive definite matrices, we can calculate a Cholesky decomposition. 
However, the dual representation has the same size as the original matrix and therefore yields no improvement. 

\begin{algorithm}
\caption{Efficient approximation of DPP sampling \label{alg:dppApprox}}
\begin{algorithmic}[1]
\Require Point set $X$, subset cardinality $k$, nearest neighbors $m$, update function $f$
\State Initialize $D = {\textbf 1}_{n}$
\For{1 to $k$}
	\State Select $i \in 1 \ldots n$ with probability $p(i) \propto D_i$ %
	\State $J \gets	J \cup i$
	\State Calculate  $\Delta_j = \| x_i - x_j \|, \  \forall j \in 1 \ldots n $ %
	\State Set $m$ nearest neighbors as neighborhood $\mN_i$ based on $\{ \Delta_j \}_{j=1 \ldots n}$
	\State $D_j \gets D_j \cdot f(\Delta_j), \ \forall j \in \mN_i$
	\State Optional: Calculate covariance $C_i$ in local neighborhood $\mN_i$ around $x_i$
\EndFor
\State \Return $J$ and optionally $\{ C_i \}_{i \in J}$
\end{algorithmic}
\end{algorithm}

To cope with the high computational costs of exact DPP sampling, we present an efficient approximation in Algorithm~\ref{alg:dppApprox}. %
The computational complexity is $\mO(ndk)$. 
Vector $D \in \mbbR^n$ models the probabilities for the selection of points as does  $\| B_i \|^2$ in the original DPP sampling algorithm. 
The algorithm proceeds by sampling $k$ points. 
Note that the cardinality of the subset cannot be set in the original DPP sampling algorithm, which led to the introduction of $k$-DPPs~\cite{kulesza2011k}. 
At each iteration we select one point $x_i$ with probability $p(i) \propto D_i$. %
Next we calculate distances $\{ \Delta_j \}_{j=1 \ldots n}$ of the selected point $x_i$ to all points in $X$. 
Based on these distances we identify a local neighborhood $\mN_i$ of $m$ nearest neighbors around the selected point $x_i$.
The update of the probabilities $D$ is restricted to the neighborhood $\mN_i$, which proves advantageous for sampling on manifolds. %
In contrast, Algorithm~\ref{alg:dpp} updates probabilities for all points. 
If we are interested in achieving a similar behavior to Algorithm~\ref{alg:dpp}, the local neighborhood should include all points.
The update function $f$ takes distances $\Delta$ as input, where we consider $f(\Delta) = \sin^2(\Delta/\tau)$ and $f(\Delta) = (1 - \exp(- \Delta^2 / 2\sigma^2))$, as motivated below. 
In subsequent iterations of the algorithm, points close $x_i$ will be selected with lower probability.

\begin{figure}[t]
\begin{center}
	\includegraphics[width=0.4\textwidth]{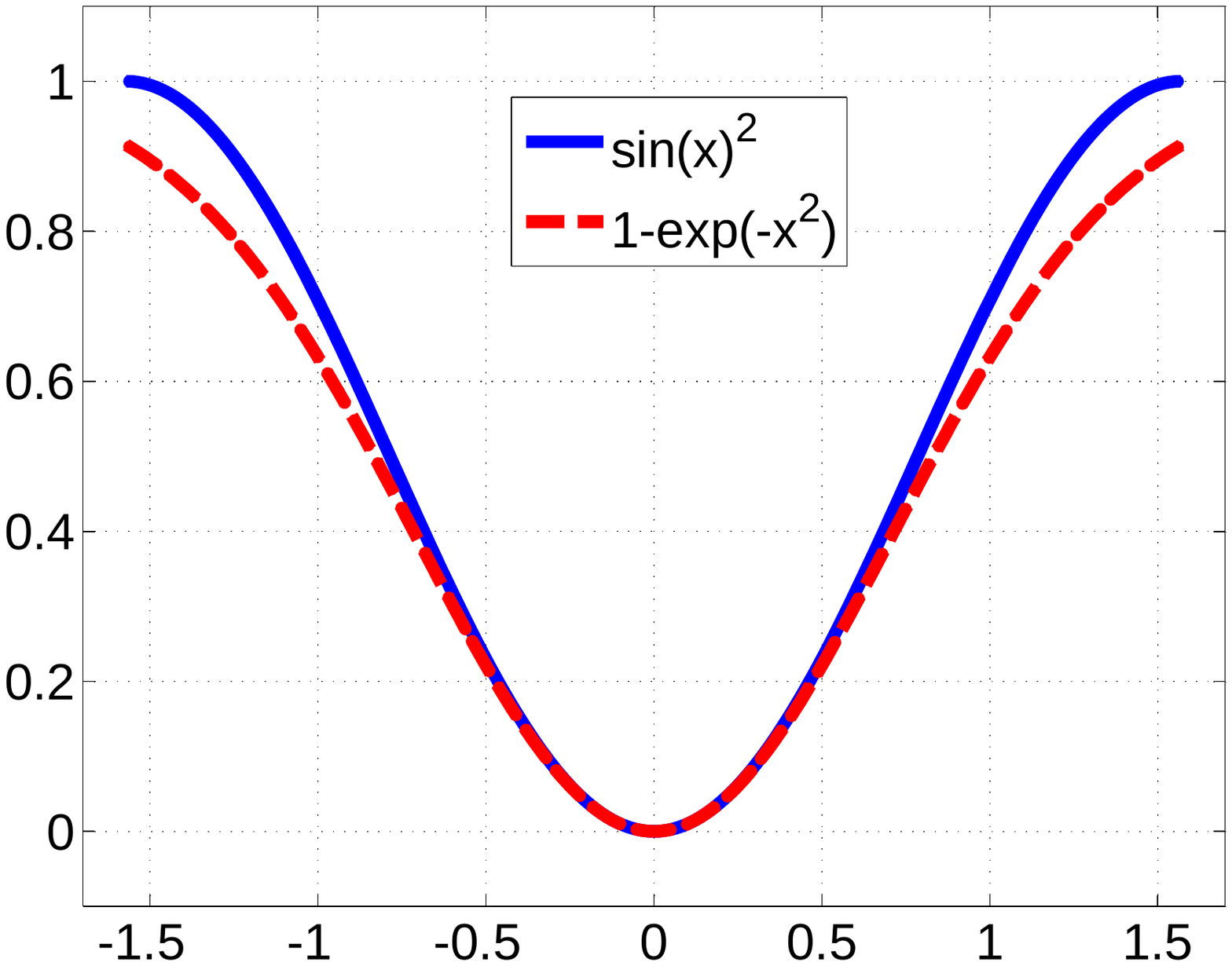} 
\caption{Comparison of $\sin^2(x)$ and $(1 - \exp(-x^2))$ used for $f$ in Algorithm 2. 
\label{fig:sine}}
\end{center}
\end{figure}

We initialize the vector $D = {\textbf 1}_{n}$, since it was noted in~\cite{kulesza2010structured} that the squared norm of the vectors $\| B_i \|^2$ gives rise initially to a fairly uniform distribution because no points have yet been selected.  
For the update step, Proposition~1 implies that the update of the probabilities for selecting a point changes with $\sin^2(\theta)$. 
The angle $\theta = \angle(B_i,B_j)$ correlates strongly with the distance $\| B_i - B_j \|$,  since $\| B_j \|$ and $\|B_i\|$ are initially the same. 
The selection of the eigenvectors in Algorithm~\ref{alg:dpp} will most likely choose the ones with the largest eigenvalues. 
We can therefore draw the analogy to multidimensional scaling (MDS)~\cite{Williams01ona} with a Gaussian kernel, where MDS selects the top eigenvectors. 
Consequently, vectors~$B_i$ correspond to low-dimensional embeddings produced by multidimensional scaling of original points~$x_i$. 
The characteristic property of MDS is to preserve pairwise distances between the original space and the embedding space, permitting the approximation $\| B_i - B_j \| \approx \| x_i - x_j\|$. 
This allows us to approximate the update dependent on the angle $\theta$ by the distance of points in the original space, $\sin^2(\theta) \approx \sin^2 ( \| x_i - x_j\| / \tau)$.
The scaling factor~$\tau$ ensures that values are in the range $[-\pi/2; \pi/2]$. 
This update is actually very similar to the Welsch function $(1 - \exp(- \| x_i - x_j\|^2 / 2\sigma^2))$, which is directly related to the weights in the kernel matrix and is commonly used in machine learning. 
We illustrate the similarity of both functions in Fig.~\ref{fig:sine} and focus on the Welsch function in our experiments. 
For subsequent iterations of the algorithm, the assumption of a similar norm of all vectors $B_i$ is violated, because the projection on the orthogonal space changes their lengths. 
Note, however, that this change is locally restricted around the currently selected point. 
Since this region is less likely to be sampled in the subsequent iterations, the assumption still holds for parts of the space that contain most probability.

\begin{figure}
\begin{center}
	\includegraphics[width=0.4\textwidth]{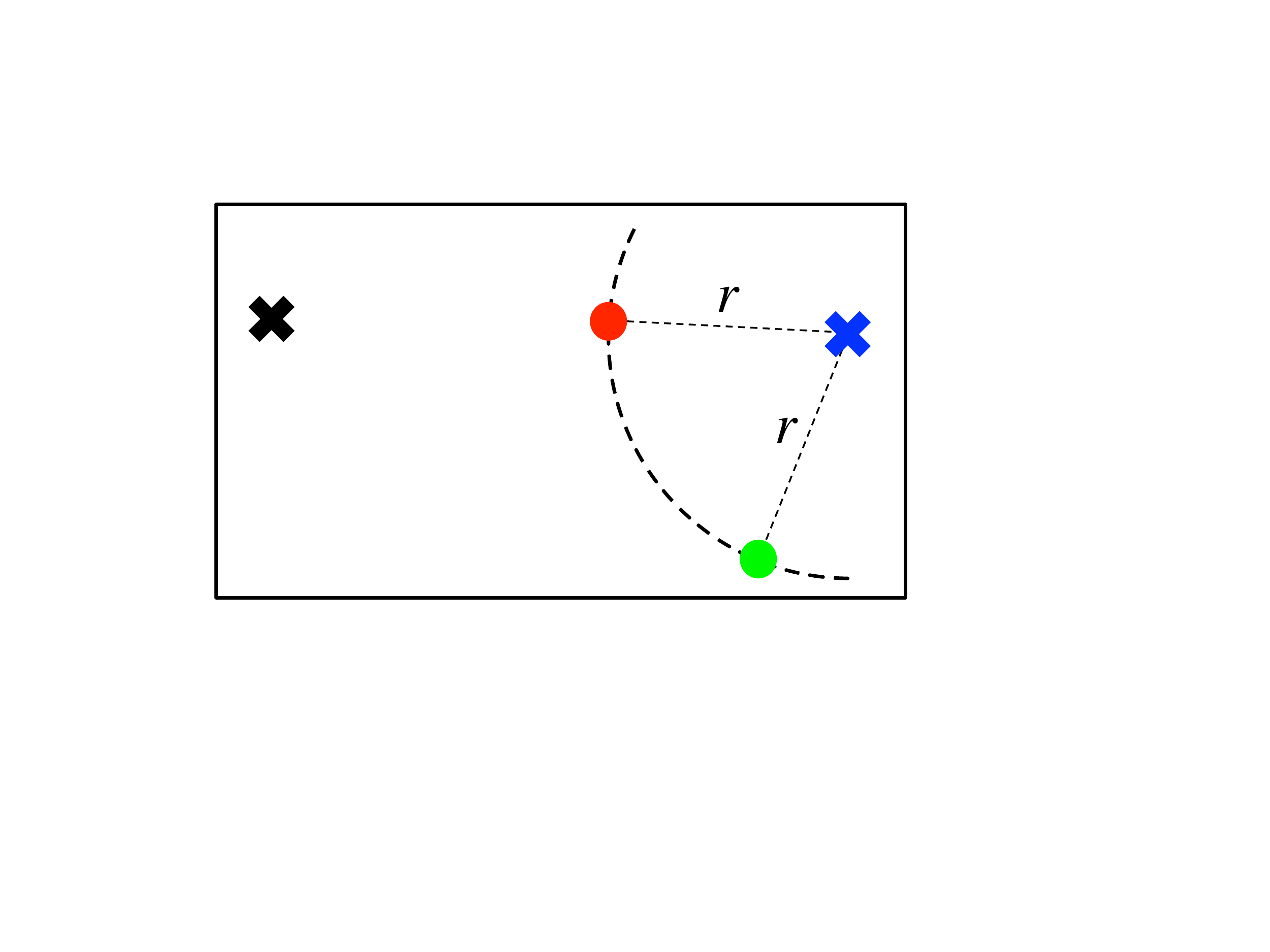} 
\caption{Two landmarks (crosses) selected in previous steps and two candidates for the current step (discs). 
K-means++ selects both candidates equally likely since they have the same distance $r$ to the landmark (blue).  %
The proposed algorithm more likely selects the green point because also the black cross influences the landmark selection, yielding an increase in diversity.}
\label{fig:kmeansPP}
\end{center}
\end{figure}

\textbf{Remark:}
The proposed algorithm bears similarities to K-means++~\cite{ArthurV07}, which replaces the initialization through uniform sampling of K-means by a new seeding algorithm. 
K-means++ seeding is a heuristic that samples points based on the minimal distance to previously selected landmarks. 
Initially, when only one landmark is selected, our proposed algorithm has a nearly identical update rule for a maximal neighborhood $\mN_i$. 
In later iterations, the algorithms differ because K-means++ bases the selection only on the distance to the nearest landmark, while all landmarks influence the probability space in our algorithm. 
Consequently, our approach potentially yields subsets with higher diversity, as illustrated in the~Fig.~\ref{fig:kmeansPP}.

\subsection{Robust Landmark-based Graph Construction}
After selecting the landmarks, the next step in the spectral analysis consists of building a graph that approximates the manifold. 
Common techniques for the graph construction include selection of nearest neighbors or $\varepsilon$-balls around each node. 
Both approaches require the setting of a parameter, either the number of neighbors or the size of the ball, which is crucial for the performance. 
Setting the parameter too low leads to a large number of disconnected components, while for many applications one is interested in having all points connected to obtain a consistent embedding of all points. 
Choosing too high values of the parameters leads to short cuts, yielding a bad approximation of the manifold. 
The appropriate selection of the parameters is more challenging on sparsely sampled manifolds. 
This is problematic for the subset selection with consecutive Nyström reconstruction because we dramatically reduce the sampling rate to limit computational complexity.

\begin{figure}[t]
\begin{center}
	\includegraphics[width=0.45\textwidth]{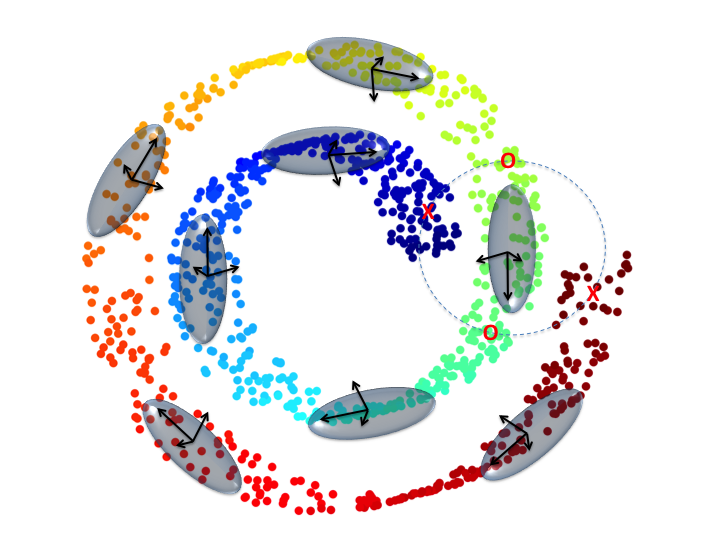} 
\caption{Nearest neighbor selection with Bhattacharyya distance. 
Covariance matrices (ellipsoids) are estimated for landmarks based on the original point set. 
Points marked as O  and X have equal Euclidean distance, easily leading to short cuts in the neighborhood graph. 
The Bhattacharyya distance considers the local geometry and is smaller for O than for X,  better approximating  the geodesic distance. 
\label{fig:covEst} }
\end{center}
\end{figure}

To address this issue, we propose a new technique for graph construction that takes the initial distribution of the points into account. 
For each landmark $x_i$, we estimate the covariance matrix $C_i$ around this point from its nearest neighbors $\mN_i$, as indicated as optional step in Algorithm~\ref{alg:dppApprox}. 
This corresponds to multivariate Gaussian distributions $\mG(x_i,C_i)$ centered at the landmark. 
A commonly used distance to compare distributions is the Bhattacharyya distance, which in case of Gaussian distributions corresponds to
\begin{equation}
B(\mG_i,\mG_j) =  \frac{1}{8} (x_i-x_j)^\top C^{-1}(x_i-x_j) + \frac{1}{2} \ln\left(\frac{|C|}{\sqrt{|C_i| |C_j|}}\right),
\label{equ:Bhat}
\end{equation}
with $C = \frac{C_i + C_j}{2}$.
This distance is less likely to produce short cuts across the manifold because points that follow the local geometry appear much closer than points that are off the geometry. 
Consequently, we replace the Euclidean distance for neighborhood selection in manifold learning with the Bhattacharyya distance, as schematically illustrated in~Fig.~\ref{fig:covEst}. 
Space requirements of this step are $\mO(d^2k)$. 
An alternative for limited space and large $d$ is to only use the diagonal entries of the covariance matrix, requiring  $\mO(dk)$. 
We summarize all steps of the proposed scalable manifold learning in Algorithm~\ref{alg:summary}.

\begin{algorithm}
\caption{Summary of scalable manifold learning \label{alg:summary}}
\begin{algorithmic}[1]
\State Select landmarks with approximate DPP sampling (Algorithm~\ref{alg:dppApprox}) 
\State Construct neighborhood graph on landmarks with Bhattacharyya distance (Equ.~\ref{equ:Bhat})
\State Calculate low-dimensional embedding based on neighborhood graph %
\State Embed non-landmark points by performing out-of-sample extension with Nystr\"om method
\end{algorithmic}
\end{algorithm}

\begin{table*}
 \centering
  \begin{tabular}{ccccccccccc}
  \toprule
  Data  & Sampling & 25 & 50 & 60 & 70 & 80 & 90 & 100 \\
  \midrule
\multirow{5}{*}{Swiss Roll} & Uniform &70.384  & 8.006  & 4.838  & 2.785  & 1.676  & 0.731  & 0.442  \\ 
 & K-means Uniform &28.124  & 3.848  & 2.319  & 1.393  & 0.756  & 0.403  & 0.235  \\ 
 & K-means++ Seeding &50.114  & 5.832  & 3.033  & 1.655  & 1.013  & 0.683  & 0.347  \\ 
 & K-means++ & \textbf{24.954}  & 3.575  & 1.915  & 1.018  & 0.711  & 0.383  & 0.222  \\ 
 & Efficient DPP &33.036  & \textbf{3.371}  & \textbf{1.466}  & \textbf{0.844}  & \textbf{0.488}  & \textbf{0.312}  & \textbf{0.202}  \\ 
 \midrule
 \multirow{5}{*}{Fish Bowl} & Uniform &44.026  & 8.394  & 6.512  & 4.678  & 1.025  & 0.935  & 0.758  \\ 
 & K-means Uniform &12.627  & 1.230  & 0.612  & 0.393  & 0.192  & 0.119  & 0.080  \\ 
 & K-means++ Seeding &22.841  & 2.337  & 1.436  & 0.643  & 0.252  & 0.089  & 0.045  \\ 
 & K-means++ &11.503  & 0.859  & 0.402  & 0.150  & 0.096  & 0.039  & 0.021  \\ 
 & Efficient DPP &\textbf{10.846}  & \textbf{0.657}  & \textbf{0.249}  & \textbf{0.095}  & \textbf{0.014}  & \textbf{0.005}  & \textbf{0.002}  \\ 
    \bottomrule
 \end{tabular}
\caption{Average reconstruction errors over 50 runs for several sampling schemes: uniform, K-means uniform, K-means++ seeding, K-means++, and efficient DPP. The subset sizes vary from 25 to 100. Best results are highlighted in bold face. }
 \label{tab:result} 
\end{table*}

\begin{figure}[t]
\begin{center}
	\includegraphics[width=0.4\textwidth]{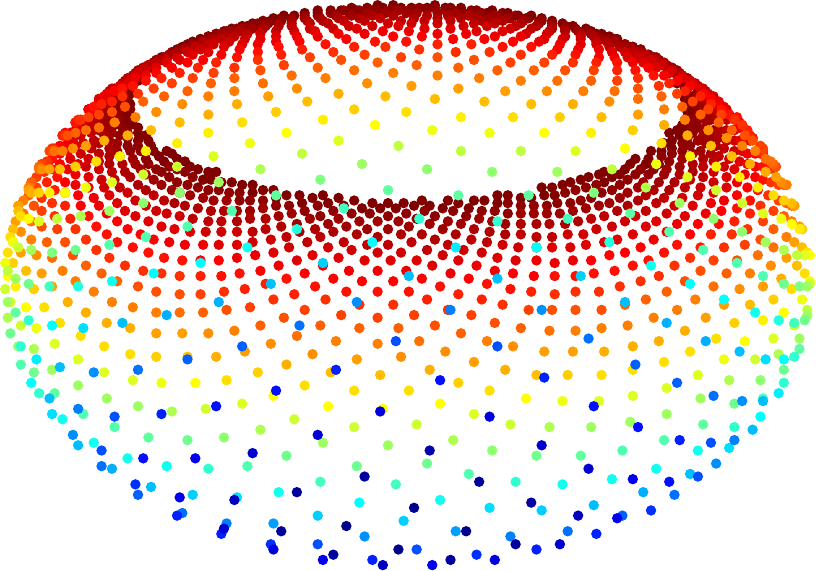} 
\caption{Fish bowl dataset used for experiments. It is a punctured sphere, which is sparsely sampled at the bottom and densely at the top. Uniform sampling is likely to pick points at the top, neglecting the bottom part. 
\label{fig:fishBowl}}
\end{center}
\end{figure}

\section{Experiments}
In our first experiment, we show that the proposed efficient DPP sampling algorithm is well suited for subset selection on non-Euclidean spaces. 
The benefit for this scenario is that we can restrict the update of the sampling probability $D$ to a local neighborhood $\mN_i$ around the current point $x_i$. 
This is in line with the motivation of many manifold learning algorithms, assuming that the space behaves locally like a Euclidean space. 
In our experiment, we set the local neighborhood $\mN_i$ to be the 20 nearest neighbors around the selected point $x_i$. 
The sampling result is shown in Fig.~\ref{fig:rollApp}. 
We obtain a point set with high diversity, covering the entire manifold. 
This illustrates that the proposed algorithm preserves the DPP characteristic on complex geometries and is therefore appropriate for subset selection in the context of non-linear dimensionality reduction.

\setcounter{subfigure}{-4}

\begin{figure*}[t]
\begin{center}
\subfigure {
	\includegraphics[width=\embeddings]{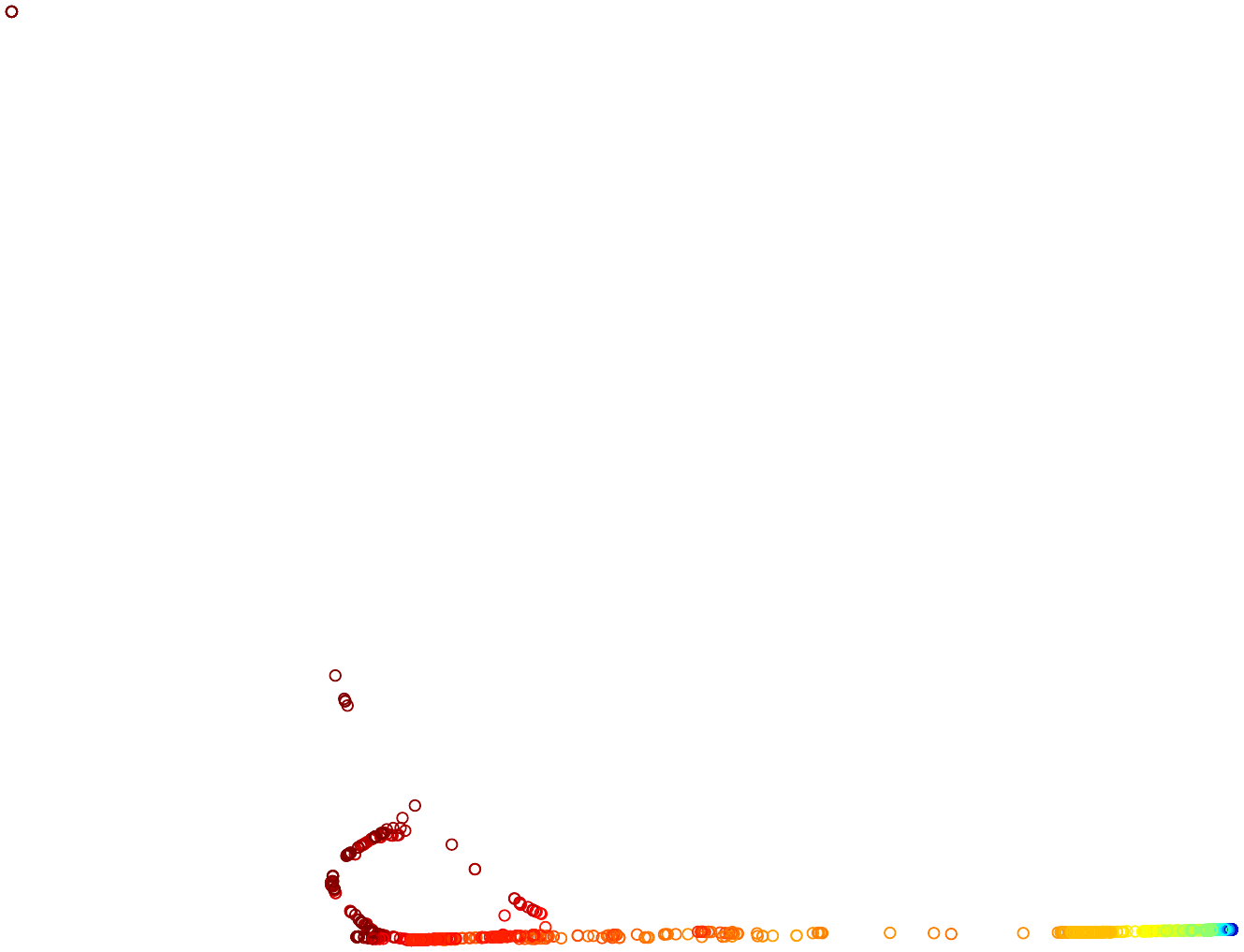} 
}
\subfigure {
	\includegraphics[width=\embeddings]{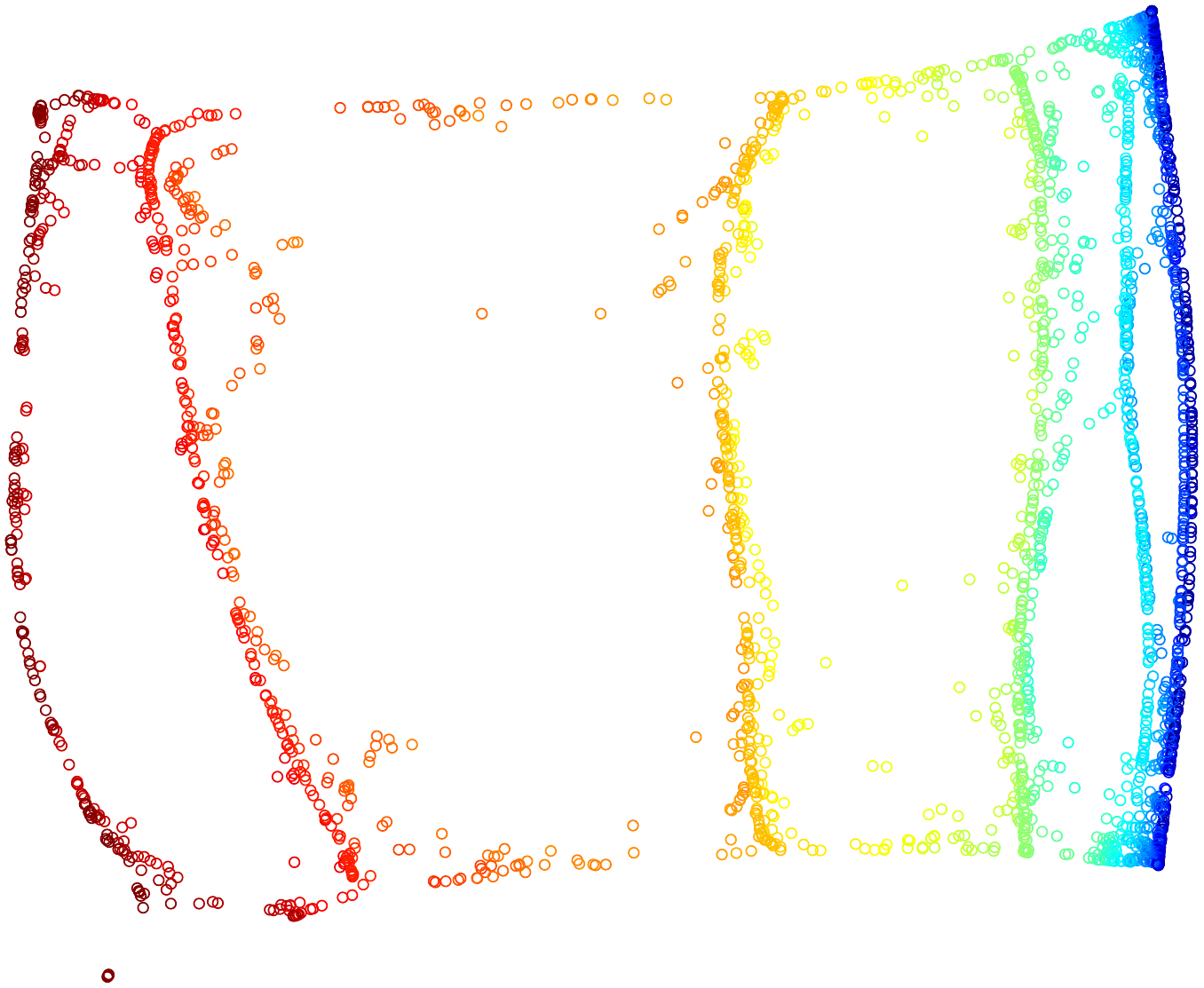} 
}
\subfigure {
	\includegraphics[width=\embeddings]{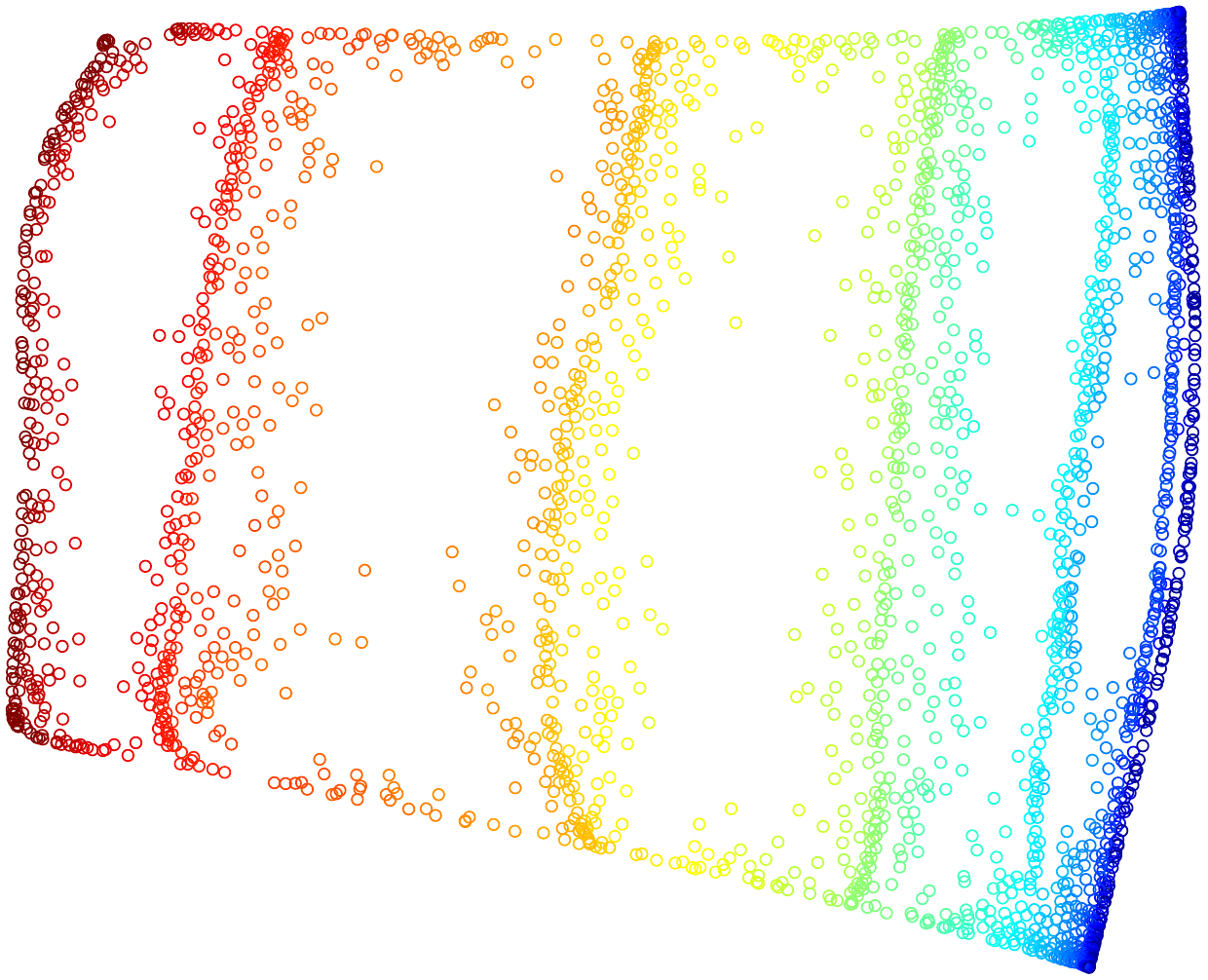} 
}
\subfigure {
	\includegraphics[width=\embeddings]{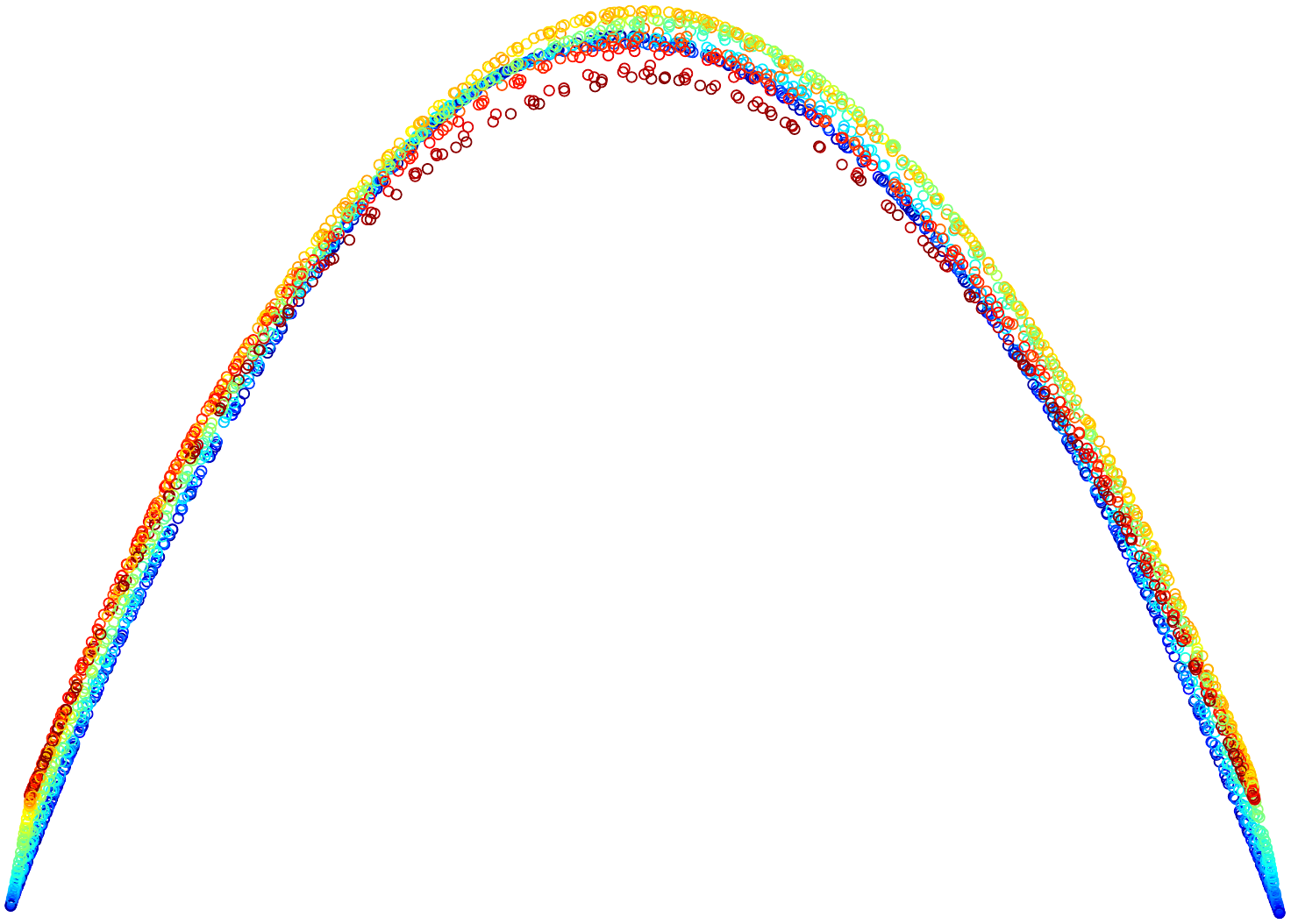} 
}
\subfigure[25] {
	\includegraphics[width=\embeddings]{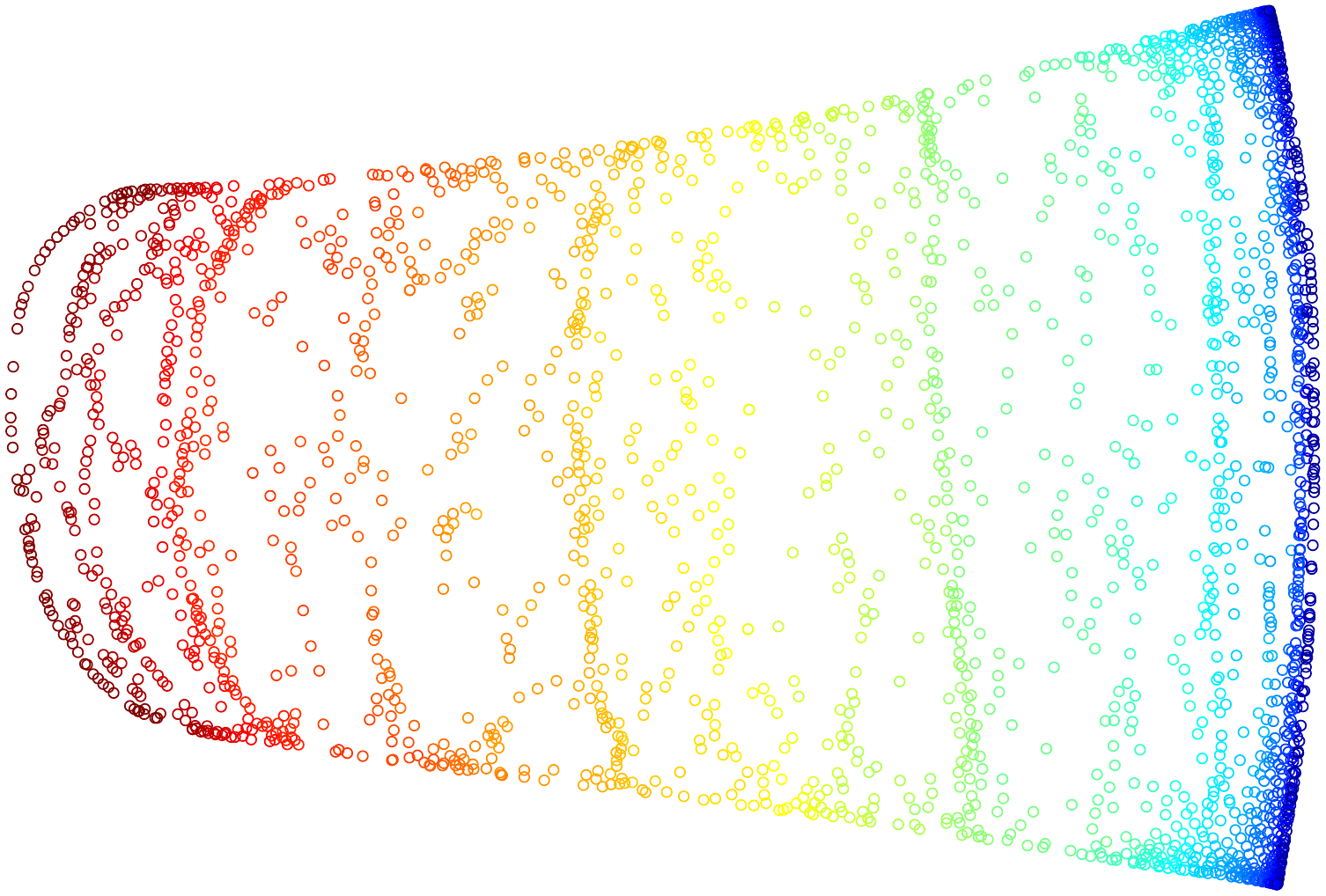} 
}
\subfigure[60] {
	\includegraphics[width=\embeddings]{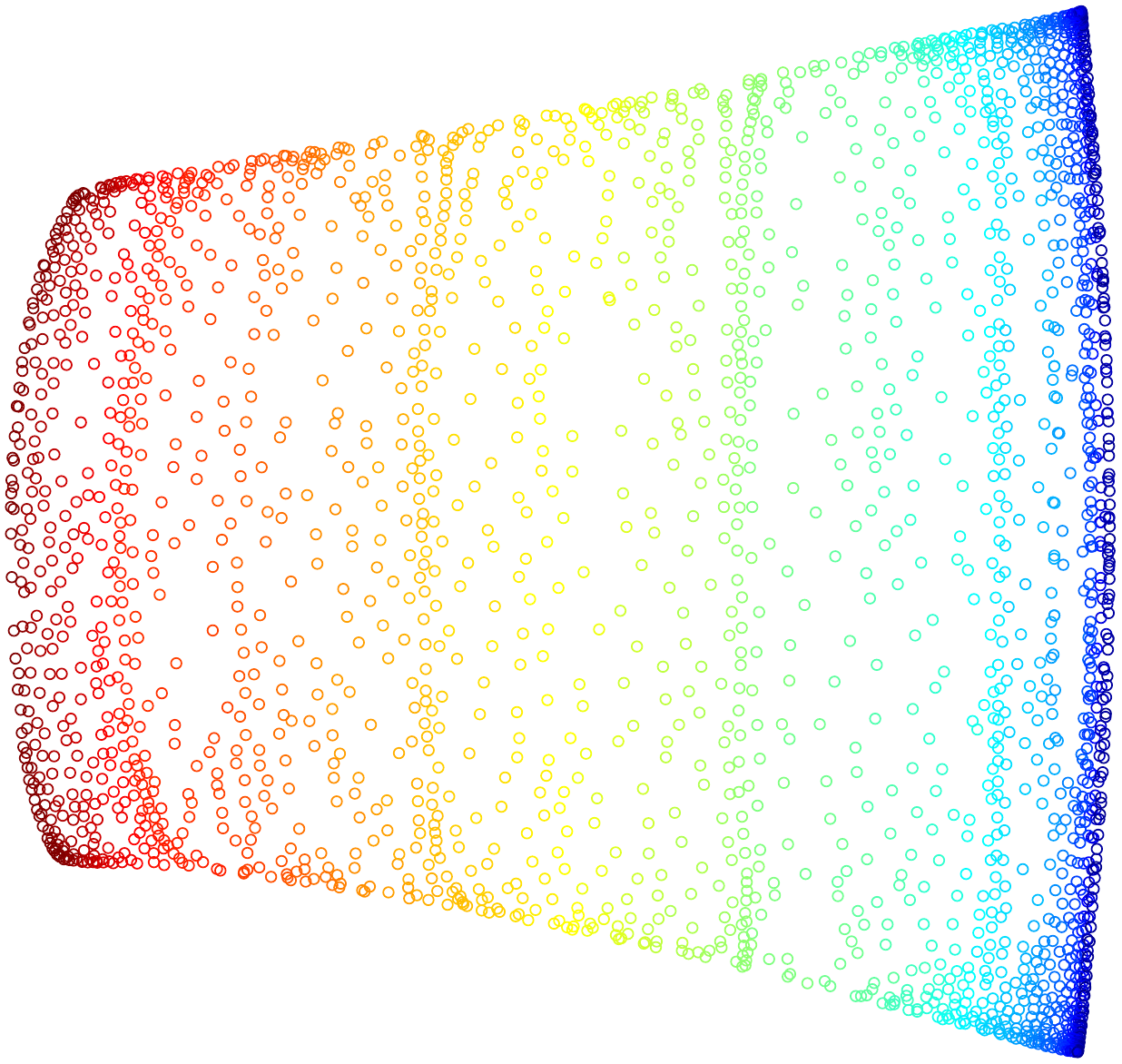} 
}
\subfigure[200] {
	\includegraphics[width=\embeddings]{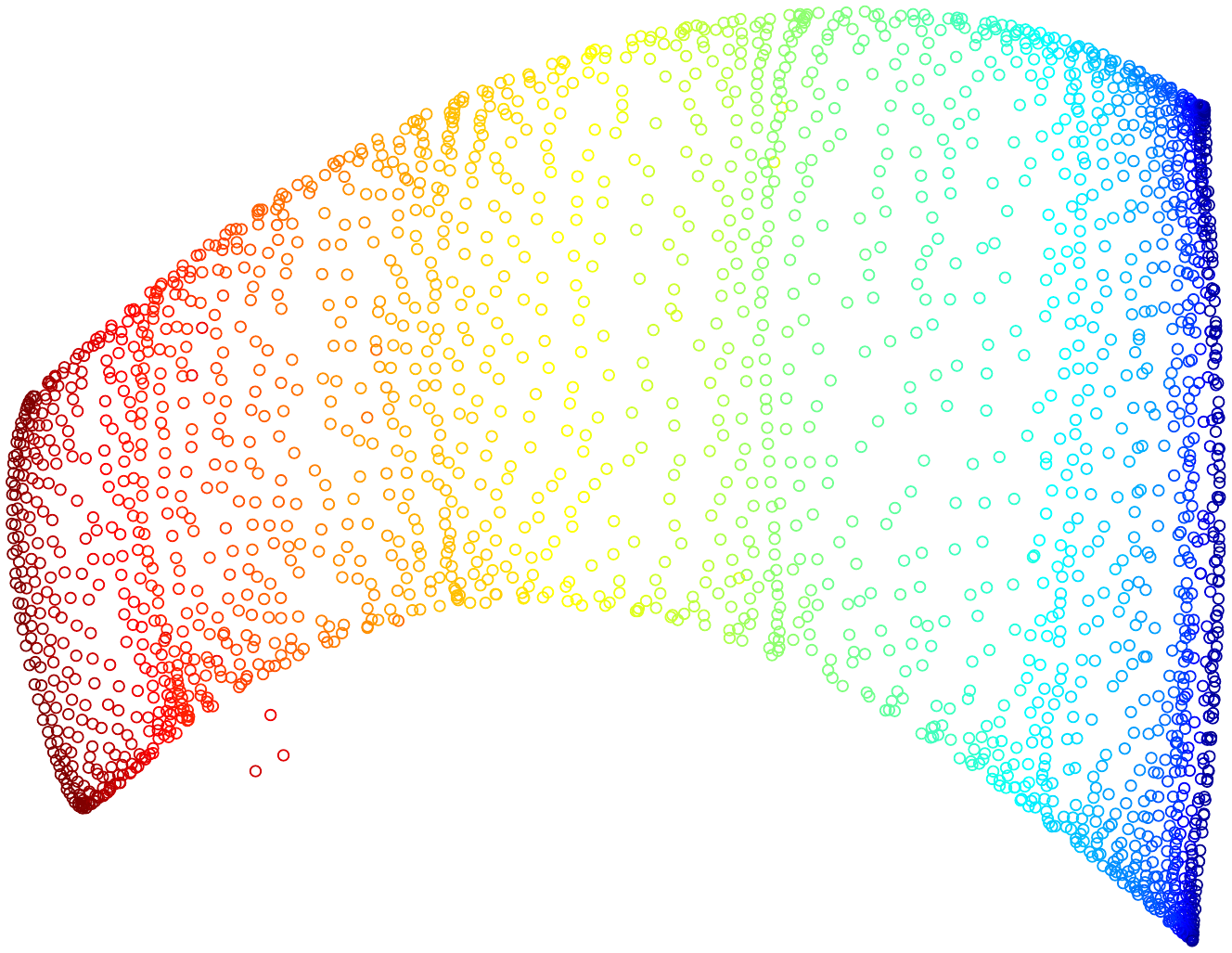} 
}
\subfigure[500] {
	\includegraphics[width=\embeddings]{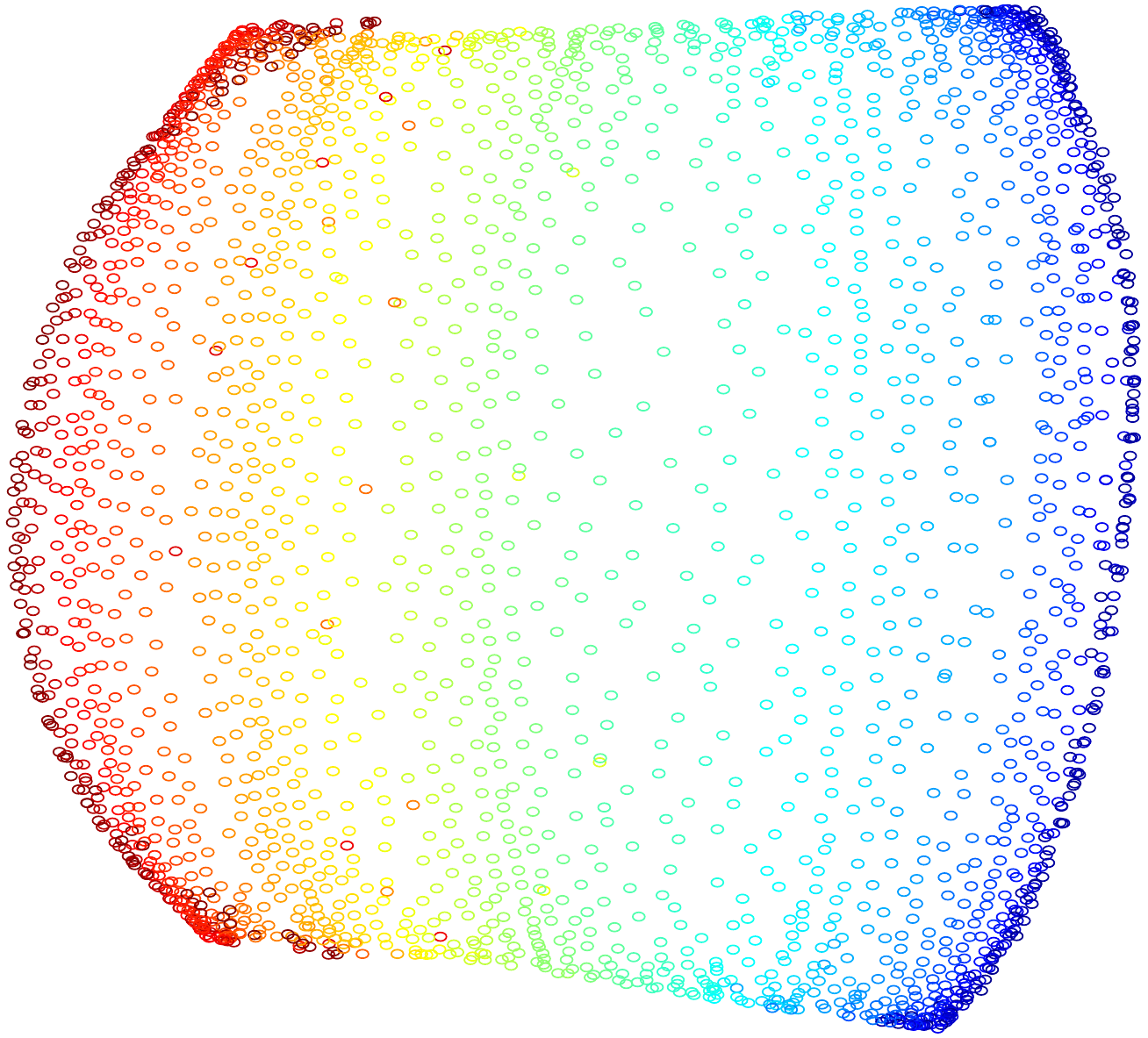} 
}
\caption{Selection of 2,500 landmarks from a set of 10 million points. 
Embedding of landmarks to 2D with Laplacian eigenmaps. 
Results for Euclidean (first row) and Bhattacharyya neighborhood selection (second row) are shown. 
Varying the number of nearest neighbors between 25 and 500, we observe that Bhattacharyya based embeddings are more robust to the parameter setting and of higher quality. 
\label{fig:embedMahal}}
\end{center}
\end{figure*}

In our second experiment, we quantify the reconstruction error for matrix completion as formulated in Equ.~(\ref{equ:error}). 
We compare the efficient DPP sampling result with uniform sampling~\cite{williams2001using} and K-means clustering with uniform seeding~\cite{zhang2008improved}, which performed best in several studies including a recent one~\cite{Kumar12}. 
Moreover, we compare to selecting the subset with the K-means++ seeding and the K-means++ algorithm, which we have not yet seen for landmark selection. 
We construct a Gaussian kernel matrix from 1,000 points on a Swiss roll (Fig.~\ref{fig:boxplot}) and on a fish bowl (Fig.~\ref{fig:fishBowl}). 
The fish bowl dataset is a punctured sphere proposed in~\cite{saul2003think}, which is sparsely sampled at the bottom and densely at the top, as shown in the supplementary material. 
We select subsets varying in size between 25 and 100 and set the parameters $\sigma = 1$ and $m = 30$ for the Swiss roll and $\sigma = 1$ and $m = 150$ for the fish bowl. 
Note that a further improvement can be achieved by adapting these parameters to the size of the subset. 
For smaller subsets, larger $\sigma$ and $m$ lead to improvements. 
We show the average reconstruction error for the different methods and datasets in Table~\ref{tab:result}, calculated over 50 different runs. 
Generally, the performance of uniform sampling is worst. The K-means++ seeding yields better results. K-means improves the results on both initializations, where K-means++ benefits from the better seeding.  
The diverse set of landmarks selected with efficient DPP sampling leads to the lowest average reconstruction errors for almost all settings. 
In our third experiment, we perform manifold learning with Laplacian eigenmaps on a point set consisting of 10 million points lying on a Swiss roll. 
The dataset is too large to apply manifold learning directly. %
We select 2,500 landmark points with the efficient DPP sampling algorithm and estimate the local covariance matrices, which we feed into the manifold learning algorithm. 
We compare the the graph construction with Euclidean and Bhattacharyya neighborhood selection. 
The graph construction yields a weight matrix $W$ and the corresponding degree matrix $\mD_{j,j} = \sum_i W_{i,j}$. 
The weight matrix $W$ is a sparse version of the kernel matrix $K_{J \times J}$, where the sparsity is controlled by the number of nearest neighbors in the graph.  
The generalized eigenvalue problem solved in Laplacian eigenmaps is 
\begin{equation}
(\mD-W) \phi_j = \lambda_j \mD \phi_j,
\end{equation}
with eigenvalues $\lambda_j$ and eigenvectors $\phi_j$. 
The $l$ eigenvectors corresponding to the $l$ smallest non-zero eigenvalues $\Lambda_{j,j} = \lambda_j, j = 1, \ldots l $ constitute the embedding in $l$-dimensional space, $\Phi_J = [\phi_1, \ldots, \phi_l]$. 
Fig.~\ref{fig:embedMahal} shows the low-dimensional embedding in 2D for Euclidean (first row) and Bhattacharyya neighborhood selection (second row). 
We vary the number of nearest neighbors in the graph from 25 to 500. 
The results show that the Bhattacharyya based neighborhood selection is much more robust with respect to the number of neighbors. 
Moreover, we notice that for Euclidean neighborhood selection the points seem to cluster along stripes. 
While we observe this effect also on the Bhattacharyya based embeddings, it is much less pronounced.

\begin{figure*}
\begin{center}
\subfigure[MNIST\label{fig:MNISTres}]{
	\includegraphics[width=0.485\textwidth]{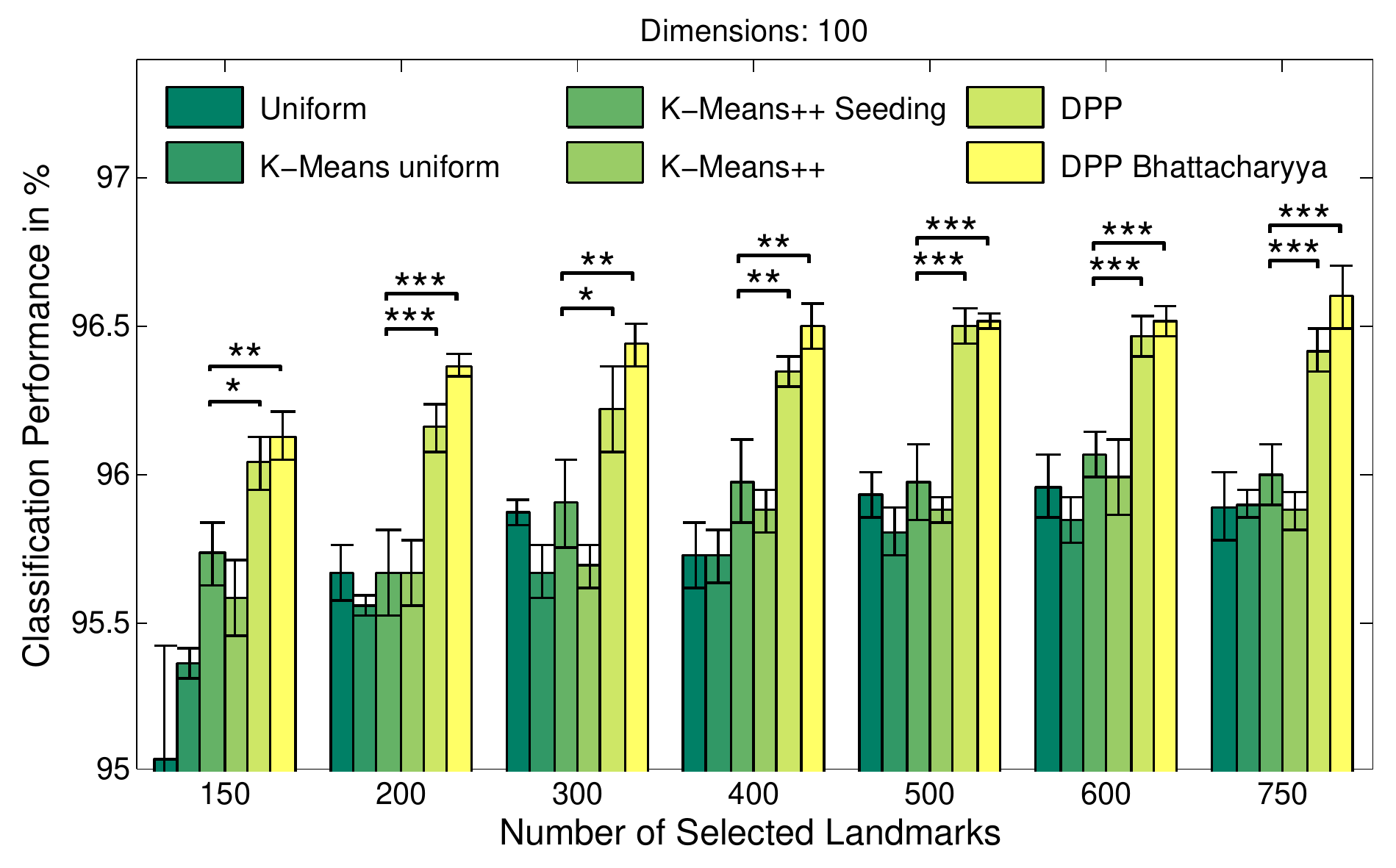}
	}
\subfigure[Head and Neck \label{fig:NeckRes}]{
	\includegraphics[width=0.48\textwidth]{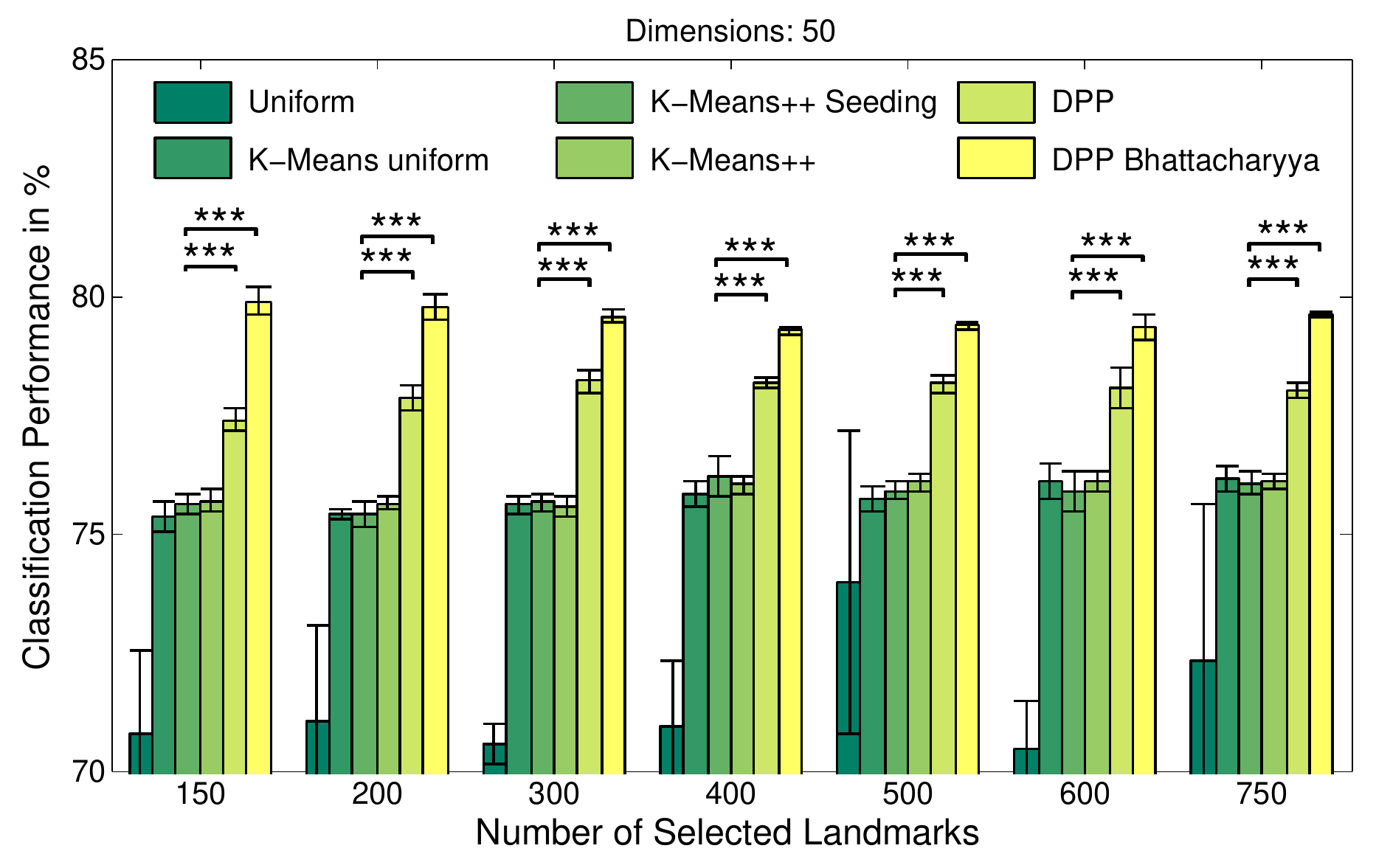}
	}	
\end{center}
\caption{Classification results for MNIST and head and neck data. %
Comparison of different subset selection schemes (uniform, K-means with uniform seeding, K-means++ seeding, K-means++, efficient DPP, efficient DPP with Bhattacharyya based graph construction) 
for varying numbers of selected landmarks. Bars indicate mean classification performance and error bars correspond to standard deviation.  
*, **, and *** indicate significance levels at $0.05$, $0.01,$ and $0.001$, respectively. 
}
\end{figure*}

\subsection{Image Data}
After having evaluated each of the steps of the proposed approach separately, we now present results for scalable manifold learning on image data. 
We work with two datasets, one consisting of handwritten digits and a second one consisting of patches extracted from 3D medical images. 
Each dataset is too large to apply manifold learning directly. 
Consequently, we select landmarks with the discussed method, perform manifold learning on the landmarks with the Bhattacharyya distance, and use the Nyström method to embed the entire point set. 
We only consider the diagonal entries of the covariance matrices due to space limitations. 
Given the low-dimensional embedding of the landmarks $\Phi_J \in \mbbR^{k \times l}$ and the diagonal matrix of eigenvalues $\Lambda \in \mbbR^{l \times l}$, we use out-of-sample extension with the Nyström method to calculate the embedding of the remaining points 
\begin{align}
\Phi_{\bar{J}} &=  \tilde{K}^\top_{J \times \bar{J}}  \Phi_{J} \Lambda^{-1}, \\
\tilde{K}_{i,j} &= \frac{K_{i,j}}{k \cdot \sqrt{ \mbbE_{i'} [K_{i',j}] \cdot \mbbE_{j'} [K_{i,j'}] }},  i \in J, j \in \bar{J} 
\end{align}
where $\tilde{K}$ is the normalized kernel for Laplacian eigenmaps and the expectation is calculated over landmark points~\cite{bengio2004out}.

Since it is difficult to quantify the quality of the embedding, we use the labels associated to the image data to perform nearest neighbor classification in the low dimensional space. 
We expect advantages for the DPP landmark selection scheme because a diverse set of landmarks spreads the entire point set in the embedding space and helps the classification. For the same reason, we also expect a good performance for the K-means++ seeding algorithm.  
Note that we abstain from pre-processing the data and from applying more advanced classifiers because we are not interested in the absolute classification performance but only in the relative performances across the different landmark selection methods.

\subsubsection{MNIST}
We work with the MNIST dataset~\cite{lecun1998gradient}, consisting of 60,000 binary images of handwritten digits for training and 10,000 for testing, with a resolution of $ 28 \times 28$ pixels. 
We set the neighborhood size to $m = 5000$ and $\sigma = 5$. 
We embed the images into 100 dimensional space with Laplacian eigenmaps. 
Fig.~\ref{fig:MNISTres} shows the statistical analysis over 20 repetitions for several landmark selection schemes across different numbers of landmarks $k$, as well as the Bhattacharyya based graph construction.  
The results show that the K-means++ seeding outperforms the uniform initialization, where K-means++ cannot further improve the initialization. 
Moreover, we observe a significant improvement in classification performance for approximate DPP sampling compared to K-Means++ seeding. Finally, the Bhattacharyya based graph construction further improves the results.

\begin{figure}
\begin{center}
\includegraphics[width=0.25\textwidth]{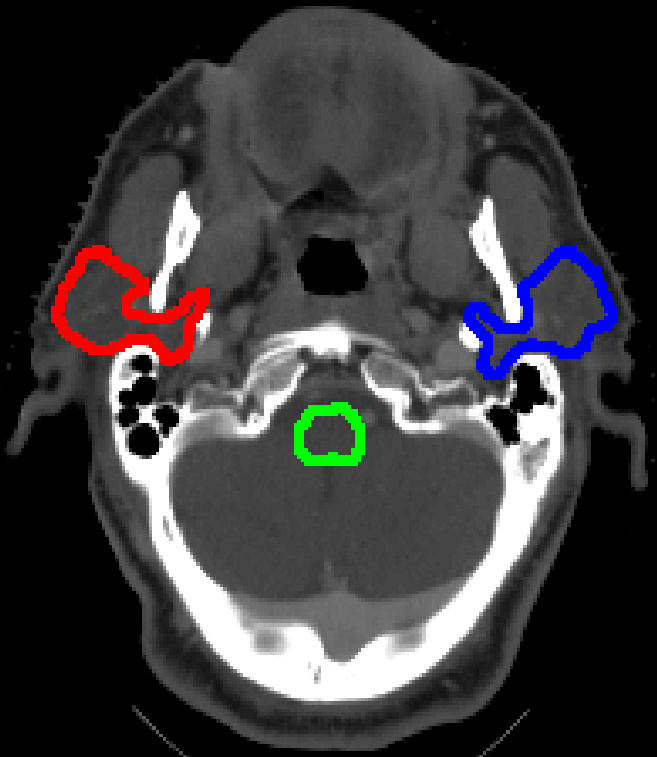}
\end{center}
\caption{Cross sectional head and neck slice. 
Overlaid are the perimeters of left parotid (green), right parotid (blue), and brainstem (green).
\label{fig:CT}}
\end{figure}

\subsubsection{Head and Neck}
In a second classification experiment, we use 3D CT scans from the head and neck region, having a resolution of $512 \times 512 \times 145$ voxels. 
These images were acquired for radiation therapy of patients with head and neck tumors. 
Fig.~\ref{fig:CT} shows one cross sectional slice with segmentations of three structures of risk: left parotid (green), right parotid (blue), and brainstem (green). 
The segmentation of these structures during treatment planning is of high clinical importance to ensure that they obtain a low radiation dose. 
We extract image patches from the left and right parotid glands, the brainstem and the surrounding region. 
We are interested in classifying patches into these four groups, where the outcome can readily serve in segmentation algorithms. 
We work with patches of size $7 \times 7 \times 3$ to reflect the physical resolution of the data which is $0.98 \times 0.98 \times 2.5$~mm$^3$. 
This results in roughly 150,000 patches extracted from three scans. 
80,000 patches are used for training and the remaining ones for testing. 
We extract landmarks from the training patches ($m = 5000$, $\sigma = 5$) and embed them into 50 dimensional space with Laplacian eigenmaps. 
Fig.~\ref{fig:NeckRes} shows the statistical analysis of the classification performance over 20 repetitions for various numbers of selected landmarks and selection schemes. 
K-means++ seeding outperforms uniform sampling. 
K-means clearly improves the uniform initialization. 
K-means++ shows a similar performance to the initial seeding. 
Similar to the experiments for MNIST, the efficient DPP approximation leads to significantly better classification results, with an additional gain for the Bhattacharyya based graph construction. 
In addition to the significant improvement, our runtime measurements showed that our unoptimized Matlab code for efficient DPP sampling runs approximately 15\% faster than an optimized MEX version of K-means.

\section{Conclusion}
We have presented contributions for two crucial issues of scalable manifold learning: 
(i) efficient sampling of diverse subsets from manifolds and (ii) robust graph construction on sparsely sampled manifolds. 
Precisely, we analyzed the sampling from determinantal distributions on non-Euclidean spaces and proposed an efficient approximation of DPP sampling. 
The algorithm is well suited for landmark selection on manifolds because probability updates are locally restricted. 
Further, we proposed the local covariance estimation around landmarks to capture the local characteristics of the space. 
This enabled a more robust graph construction with the Bhattacharyya distance and yielded low dimensional embeddings of higher quality. 
We compared to state-of-the-art subset selection procedures and obtained significantly better results with the proposed algorithm. 

\noindent
\textbf{Acknowledgements:} This work was supported in part by the Humboldt foundation, the National Alliance for Medical Image Computing (U54-EB005149), and the NeuroImaging Analysis Center (P41-EB015902).

{\small
\bibliographystyle{splncs03}
\bibliography{jab_bib}
}

\end{document}